\documentclass[journal]{IEEEtran}
\usepackage[switch]{lineno}
\usepackage{amsmath}
\usepackage{amssymb}
\usepackage{subfigure}
\usepackage{multirow}
\usepackage{multicol}
\usepackage[colorlinks=false]{hyperref}
\usepackage{graphicx}
\usepackage{booktabs}
\usepackage{color}

\begin{document}
\title{Learning Efficient GANs for Image Translation via \\Differentiable Masks and co-Attention Distillation}

\author{Shaojie Li,
        Mingbao Lin,
        Yan Wang, 
        Fei Chao,~\IEEEmembership{Member,~IEEE},
        Ling Shao,~\IEEEmembership{Fellow,~IEEE},
        Rongrong Ji,~\IEEEmembership{Senior Member,~IEEE}% <-this % stops a space
\IEEEcompsocitemizethanks{
\IEEEcompsocthanksitem S. Li and F. Chao are with the Media Analytics and Computing Laboratory, Department of Artificial Intelligence, School of Informatics, Xiamen University, Xiamen 361005, China (e-mail: rrji@xmu.edu.cn).\protect
\IEEEcompsocthanksitem M. Lin is with the Media Analytics and Computing Laboratory, Department of Artificial Intelligence, School of Informatics, Xiamen University, Xiamen 361005, China, also with the Youtu Laboratory, Tencent, Shanghai 200233, China.
\IEEEcompsocthanksitem Y. Wang is with Pinterest, USA.\protect
\IEEEcompsocthanksitem L. Shao is with the Inception Institute of Artificial Intelligence, Abu Dhabi, United Arab Emirates, and also with the Mohamed bin Zayed University of Artificial Intelligence, Abu Dhabi, United Arab Emirates.
\IEEEcompsocthanksitem R. Ji (Corresponding Author) is with the Media Analytics and Computing Laboratory, Department of Artificial Intelligence, School of Informatics, Xiamen University, Xiamen 361005, China, also with Institute of Artificial Intelligence, Xiamen University, Xiamen 361005, China (e-mail: rrji@xmu.edu.cn).\protect
}% <-this % stops an unwanted space
\thanks{Manuscript received April 19, 2005; revised August 26, 2015.}}

\markboth{IEEE TRANSACTIONS ON MULTIMEDIA}%
{Shell \MakeLowercase{\textit{et al.}}: Bare Demo of IEEEtran.cls for IEEE Journals}

\maketitle

\begin{abstract}
Generative Adversarial Networks (GANs) have been widely-used in image translation, but their high computation and storage costs impede the deployment on mobile devices. Prevalent methods for CNN compression cannot be directly applied to GANs due to the peculiarties of GAN tasks and the unstable adversarial training. To solve these, in this paper, we introduce a novel GAN compression method, termed DMAD, by proposing a Differentiable Mask and a co-Attention Distillation. \textit{The former} searches for a light-weight generator architecture in a training-adaptive manner. To overcome channel inconsistency when pruning the residual connections, an adaptive cross-block group sparsity is further incorporated. \textit{The latter} simultaneously distills informative attention maps from both the generator and discriminator of a pre-trained model to the searched generator, effectively stabilizing the adversarial training of our light-weight model. Experiments show that DMAD can reduce the Multiply Accumulate Operations (MACs) of CycleGAN by 13$\times$ and that of Pix2Pix by 4$\times$ while retaining a comparable performance against the full model.  Our code can be available at \url{https://github.com/SJLeo/DMAD}.
\end{abstract}

\begin{IEEEkeywords}
Generative Adversarial Networks, GAN Compression, Network Pruning, Knowledge Distillation, Image Translation.
\end{IEEEkeywords}

\IEEEpeerreviewmaketitle

\section{Introduction}
\begin{figure}
\begin{center}
\includegraphics[height=0.3\linewidth]{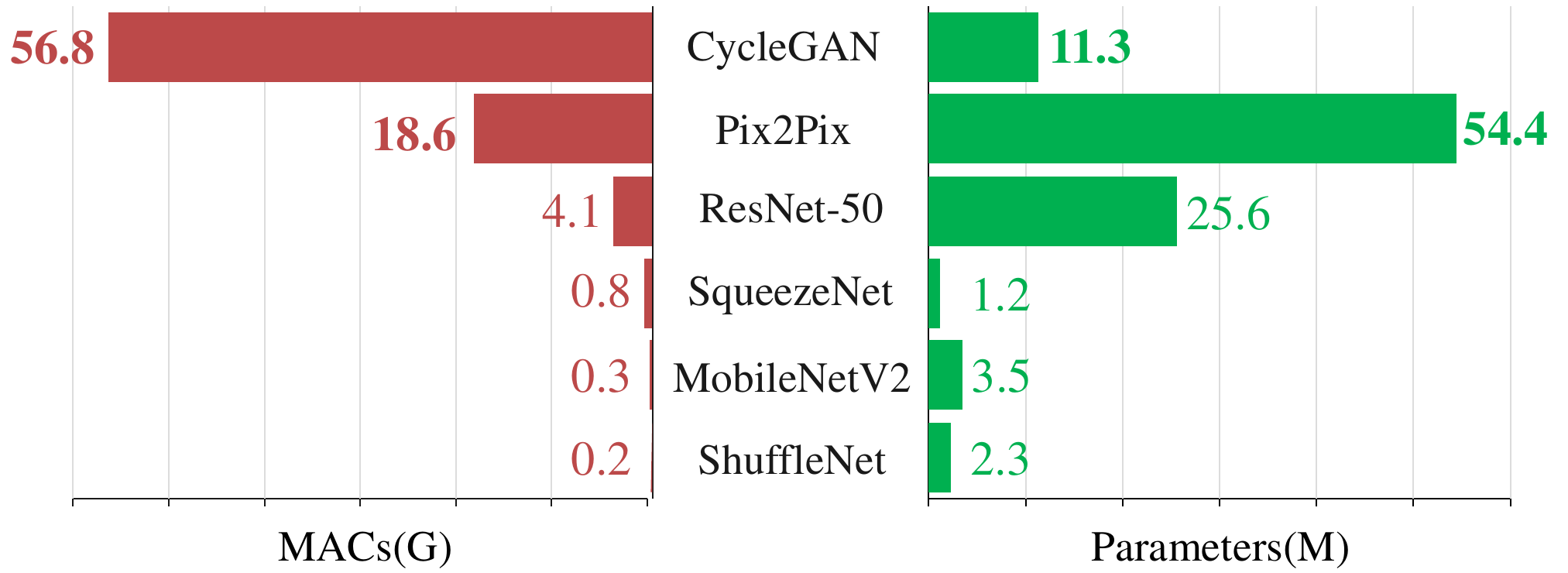}
\end{center}
\vspace{-1em}
\caption{Comparison on MACs and parameters between GANs including CycleGAN~\cite{zhu2017unpaired} and Pix2Pix~\cite{isola2017image}, and traditional CNNs including ResNet-50~\cite{he2016deep}, SqueezeNet~\cite{iandola2016squeezenet}, MobileNetV2~\cite{sandler2018mobilenetv2} and ShuffleNet~\cite{zhang2018shufflenet}. The GANs are much more complex than the traditional CNNs.}
\label{macs_params_compare}
\vspace{-1em}
\end{figure}

\IEEEPARstart{G}{enerative} Adversarial Networks (GANs)~\cite{goodfellow2014generative} have led many vision tasks, such as image synthesis~\cite{brock2018large}, domain translation~\cite{isola2017image}, image-to-image translation~\cite{zhu2017unpaired}, \emph{etc}. However, it comes with great costs on computation and memory. For example, from Fig.\ref{macs_params_compare}, the popular CycleGAN~\cite{zhu2017unpaired} requires over 56.8G MACs (Multiply-Accumulate Operations) for a single $256 \times 256$ image, which is 13$\times$ ResNet-50~\cite{he2016deep} and 189$\times$ MobileNetV2~\cite{sandler2018mobilenetv2}. Besides, the Pix2Pix~\cite{isola2017image} also disadvantages in its parameter number, which is 2$\times$ ResNet-50 and 15$\times$ MobileNetV2. The huge resource demand makes their deployment on mobile devices impractical. Thus, learning efficient GANs has become an important task.

To overcome this, great effort has been made to reduce general neural network complexity, including network pruning~\cite{ding2019global,wang2020pruning}, weight quantization~\cite{leng2018extremely,zhou2018adaptive}, tensor decomposition~\cite{denil2013predicting,hayashi2019exploring}, and knowledge distillation~\cite{hinton2015distilling,zagoruyko2017paying}. 

\begin{figure*}[!t]
\begin{center}
\includegraphics[width=0.65\linewidth]{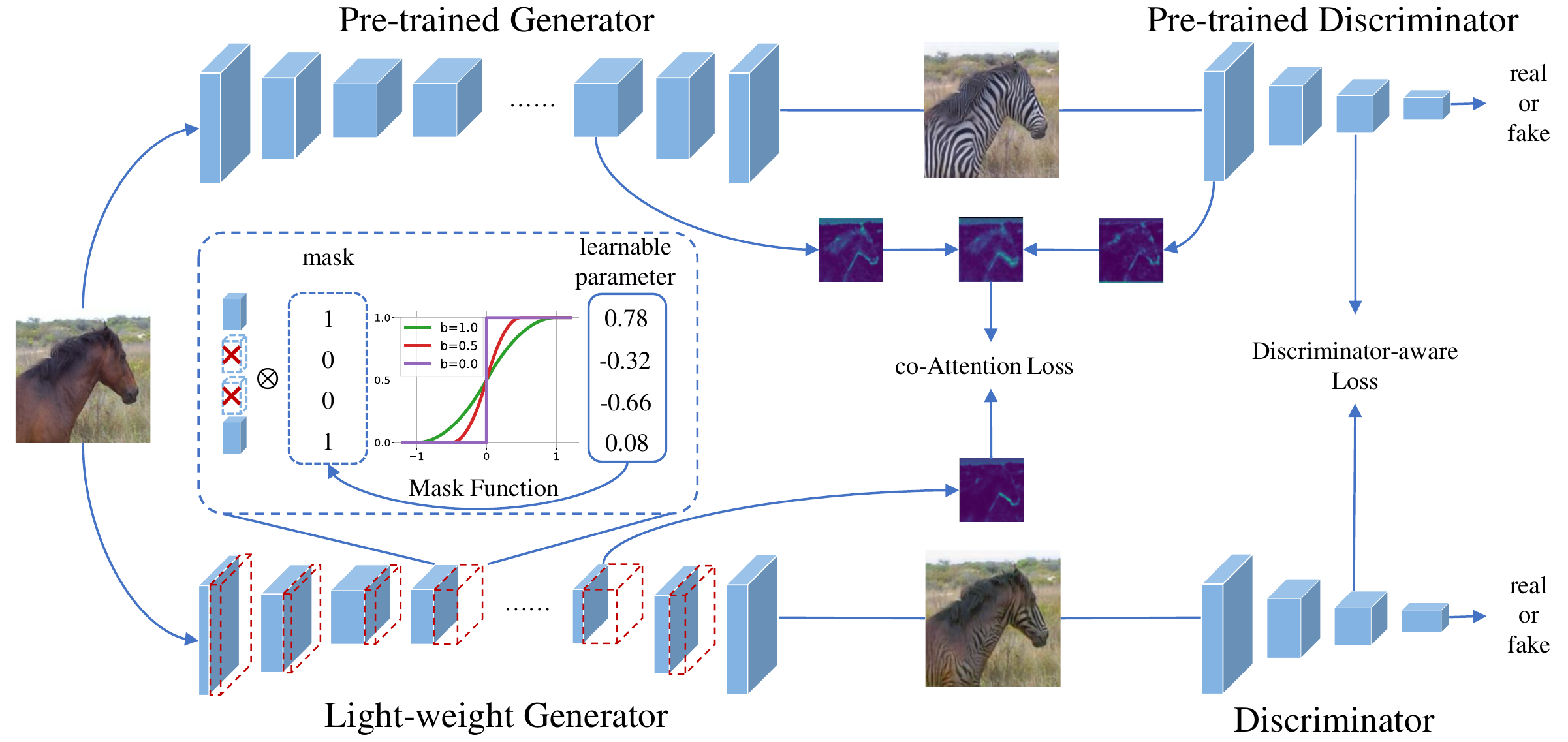}
\end{center}
\vspace{-1em}
\caption{Framework of our DMAD. We first build a pre-trained model similar to a GAN network, upon which a differentiable mask is imposed to scale the convolutional outputs of the generator and derive a light-weight one. Then, the co-Attention of the pre-trained GAN and the outputs of the last-layer convolutions of the discriminator are distilled to stabilize the training of the light-weight model.} 
\label{framework}
\vspace{-1em}
\end{figure*}

While great progress has been made, the above models are specially designed for CNNs.
They cannot be directly applied to GANs because of two reasons. 
First, different from CNNs, the generators in GANs require significantly more parameters and more complicated computation to construct high-dimensional mapping, such as image-to-image translation, style transfer, \emph{etc}. In addition, GAN generation tasks often lack ground-truth class labels. These peculiarities of GAN tasks make compression more challenging than other vision tasks such as classification and detection. Thus many handcrafted pruning criteria for CNNs are not compatible with GANs, and directly extending them to GANs has been demonstrated inadvisable~\cite{shu2019co,fu2020autogan}. 
Second, GANs also differ from CNNs in their notorious instability of adversarial training, where the generator and discriminator are alternately updated as a two-player competition.
% Finally, most CNN distillation approaches only consider the final label as supervision, and thus overlook the quality and diversity of the generated images.
% the quality of the image generated by the generator only depends on the discriminator and does not have exact ground-truth, the diversity of generation tasks should be considered in the knowledge distillation process.
% Considering the diversity of the generated images, it not wise to directly imitate the intermediate feature map or output of the pre-trained generator. All of these make it hard to directly extend conventional CNN compression to GANs. 
Therefore, a GAN-specific distillation framework is necessary for effective compression.

% Furthermore, the direction of distillation may be challenging for GANs. First, the generators in GANs require significantly larger parameters and heavier computation to match classless tasks such as image-to-image translation, style transfer, \emph{etc}, making it hard to directly extend conventional CNN compression to GANs. Second, GANs also differ from CNNs in their adversarial training, where the generator and discriminator are alternately updated as a two-player competition, which is unstable. Therefore, training student GANs is more challenging than CNNs. \yan{Better to check whether my changes reflect your intention.}

%
In this paper, we propose a novel GAN compression method, termed DMAD, by introducing a Differentiable Mask and co-Attention Distillation, as illustrated in Fig.\ref{framework}. We design a differentiable mask to carry out architecture search in a training-adaptive manner to get rid of the time-consuming NAS process and handcrafted pruning criteria. At the same time, we use the attention maps of both the pre-trained generator and discriminator to stabilize the training of the light-weight generator and avoid excessive impact on the diversity of the generated results. Specifically, the convolutional outputs in the given generator network are scaled by a mask function which is differentiable in relation to its learnable input. Starting with a smooth and gentle shape, our mask can adaptively degenerate to the standard step function, in which filters with zero masks will be removed. Thus the light-weight generator can be derived automatically. %, which not only gets rid of the heavy search process of NAS, but also does not depend on specific network structures. Besides, to solve the channel inconsistency when pruning networks across residual connections \yan{Double check my change makes sense}, an adaptive cross-block group sparsity constraint is further proposed, where filters from different blocks but with the same indices are collaboratively forced towards zero masks so that they can be safely removed together. 
However, the light-weight generator has limited performance using traditional adversarial training. Therefore we introduce a co-attention loss to provide effective supervision for a stable training process. The loss considers the attention maps of both the pre-trained generator and discriminator. With the feature maps from the last layer of the pre-trained discriminator, the co-attention loss guides the distillation of the light-weight generator and discriminator to boost the quality of generated images.

\section{Related Work}

Generative Adversarial Networks (GANs)~\cite{goodfellow2014generative} are good at synthesizing realistic results.
Among them, Image-to-Image GANs has been widely applied in image translation~\cite{isola2017image, emami2020spa}, image editing~\cite{he2019attgan, karras2020analyzing}, style transfer~\cite{zhu2017unpaired, azadi2018multi}, image super-resolution~\cite{zhang2020supervised, ledig2017photo, guo2019auto}, face processing~\cite{ngo2021self, peng2019cgr}.
Isola et al.~\cite{isola2017image} propose Pix2Pix to solve a wide range of supervised image translation tasks. 
In addition to adversarial training loss, Pix2Pix leverages pixel-wise regression loss between the generated image and the ground truth to ensure the quality of the translated images.
Zhu et al.~\cite{zhu2017unpaired} propose CycleGAN for unsupervised image translation.
CycleGAN trains two cross-domain transfer GANs with well-designed cyclic losses.
However, as shown in Fig.~\ref{macs_params_compare}, Pix2Pix and CycleGAN require a lot of computations and memory overheads.
Great progress has been made to reduce the complexity of neural networks.
Among them, network pruning and knowledge distillation are effective methods to reduce network complexity.

Network pruning achieves model compression by removing redundant operations in the network. Many works focus on removing individual neurons according to a given criterion, \emph{e.g.}, second-order Taylor expansion~\cite{lecun1990optimal}, $\ell_2$-norm regularization~\cite{han2015learning}, global sparse momentum SGD~\cite{ding2019global}, \emph{etc}. Another group of methods remove the whole filters directly~\cite{liu2017learning, zhuang2018discrimination,lin2020hrank, wang2020pruning}. For example, network slimming~\cite{liu2017learning} imposes a sparsity on the scaling factor of batch normalization and removes filters with smaller scaling values. Discrimination-aware channel pruning~\cite{zhuang2018discrimination} uses the output of the intermediate layer for pruning. HRank~\cite{lin2020hrank} prunes filters with lower-rank feature map outputs. The study~\cite{wang2020pruning} explores pruned structures by pruning from randomly initialized weights.

In addition to pruning, knowledge distillation~\cite{hinton2015distilling} has also attracted the community's attention. By transferring knowledge from a larger teacher network, the learning ability of a smaller student network can be enhanced. To that effect, The works~\cite{romero2015fitnets, zagoruyko2017paying, yim2017gift} make full use of the intermediate representations from the teacher network to improve the performance of a student network. Rather than directly using the intermediate representations, Zagoruyko et al.~\cite{zagoruyko2017paying} propose to utilize spatial attention as a mechanism of knowledge transfer. Yim et al.~\cite{yim2017gift} distill the second-order statistics of a Gram matrix to complete the teaching. In the work~\cite{chung2020feature}, feature-map-level online distillation is proposed, and then implemented in an adversarial training manner. 

Since recent two years, related works in compressing GANs have emerged. Shu et al.~\cite{shu2019co} propose to compress CycleGAN for unpaired image translation with the cycle consistency loss as the co-evolution fitness to update a sub-structure set. However, this strategy is very limited to networks with a cycle loss and cannot be straightforwardly extended to other GANs, such as Pix2Pix. 
% Chen et al.~\cite{chen2020distilling} propose to compress GANs by distilling both the generator and discriminator information of a pre-trained model, while using a handcrafted student generator. 
Distillation-based methods~\cite{aguinaldo2019compressing, chen2020distilling, li2020semantic, tsunashima2021adversarial} propose to compress GANs by distilling information of a pre-trained model to a hand-crafted student generator.
With the advance of network architecture search (NAS), recent works~\cite{fu2020autogan,li2020gan} have used this to automatically derive light-weight generator models. However, this inevitably introduces high search complexity. The search space in~\cite{fu2020autogan} is very broad, including operations and widths that have to be pre-defined manually. Though Li et al.~\cite{li2020gan} incorporate once-for-all training to model a super-net and thus reduce the search complexity, the sub-net combination is still huge and results in a great burden on training. 
% Wang et al.~\cite{wang2020gan} directly use pruning method~\cite{liu2017learning} to greatly simplify the compression process, but the compression effect is not ideal.
The compression methods~\cite{wang2020gan, gong2021towards, jin2021teachers} combine pruning and knowledge distillation to compress GANs, but the compression effectiveness is still limited. Recent studies~\cite{chen2021gans, chen2021data} utilize lottery ticket theory~\cite{frankle2019lottery} to obtain an unstructured sparse generator. However, it requires special hardware to achieve practical acceleration.
%Overall, the above methods are either trapped in the tedious and time-consuming NAS search, or the manual design of the network which lacks flexibility.

\section{Methodology}

GANs consist of a generator G and a discriminator D. Considering the source image $\mathbf{X} = \{ \mathbf{x}_1, \mathbf{x}_2, ..., \mathbf{x}_n \}$ and target image $\mathbf{Y} = \{ \mathbf{y}_1, \mathbf{y}_2, ..., \mathbf{y}_n \}$, GANs optimization is given as:
\begin{equation}
\begin{split}
\mathcal{L}_{\text{GAN}}=&\mathbb{E}_{\mathbf{x},\mathbf{y}}[\log D (\mathbf{x},\mathbf{y})] + \mathbb{E}_{\mathbf{x}}[\log(1-D (\mathbf{x}, G(\mathbf{x})))].
\end{split}
\end{equation}

The generator aims to map the source image $\mathbf{x}$ to the target image $\mathbf{y}$ to cheat the discriminator, while the discriminator aims to tell the real image $\mathbf{y}$ from the generated $G(\mathbf{x})$. Besides, for image translation, additional loss $\mathcal{L}_{\text{SPE}}$ is designed for the generator to complete a specific task, \emph{e.g.}, the cycle consistency loss $\mathcal{L}_{\text{SPE}}=\mathbb{E}_{\mathbf{x}}[\|G_2(G_1(\mathbf{x}))-\mathbf{x}\|_2]$ in CycleGAN, and the pixel-wise regression loss $\mathcal{L}_{\text{SPE}}=\mathbb{E}_{\mathbf{x},\mathbf{y}}[\|y-G(\mathbf{x})\|_1]$ in Pix2Pix. Thus, the objective for image translation is:
\begin{equation} 
\mathcal{L}_{\text{cGAN}}=\mathcal{L}_{\text{GAN}}+
\lambda_{\text{SPE}} \mathcal{L}_{\text{SPE}},
\label{cGAN_objective}
\end{equation}
where $\lambda_{\text{SPE}}$ is a pre-given hyperparameter.

\subsection{Preliminaries}

Consider a pre-trained GAN with a generator $G_T$ and discriminator $D_T$. The generator $G_T$ consists of $L$ convolutional layers and we denote its filter weights as $\mathbf{W}^{G_T} = \{\mathbf{W}_i^{G_T}\}_{i=1}^L$. The $i$-th filter weight is given by $\mathbf{W}_i^{G_T} \in \mathbb{R}^{n_i \times c_i \times h_i \times w_i}$, where $n_i, c_i, h_i$ and $w_i$ denote the filter number, channel number, filter height and width of the $i$-th layer. Similarly, we have the filter weights for the discriminator $\mathbf{W}^{D_T} = \{\mathbf{W}_i^{D_T}\}_{i=1}^{L'}$.

Our goal is to derive a light-weight generator $G_S$, filter weights of which are denoted as $\tilde{\mathbf{W}}^{G_S} = \{\tilde{\mathbf{W}}_i^{G_S}\}_{i=1}^L$ with the $i$-th layer weight $\tilde{\mathbf{W}}_i^{G_S} \in \mathbb{R}^{\tilde{n}_i \times \tilde{c}_i \times h_i \times w_i}$ under strict constraints of $\tilde{n}_i \le n_i$ and $\tilde{c}_i \le c_i$. The set $\{\tilde{n}_i\}_{i=1}^L$ constitutes the network architecture of the light-weight generator $G_S$, which should have comparable performance against the pre-trained $G_T$ after end-to-end adversarial training.

Thus, deriving an outstanding $G_S$ lies in: how to obtain the network architecture effectively and efficiently, and how to stabilize the adversarial training to recover the light model performance. In what follows, we devise a differentiable mask and a co-attention distillation to solve the above problems.

\subsection{Differentiable Mask}

\begin{figure}
\centering
\subfigure[forward]{\includegraphics[width=0.4\columnwidth]{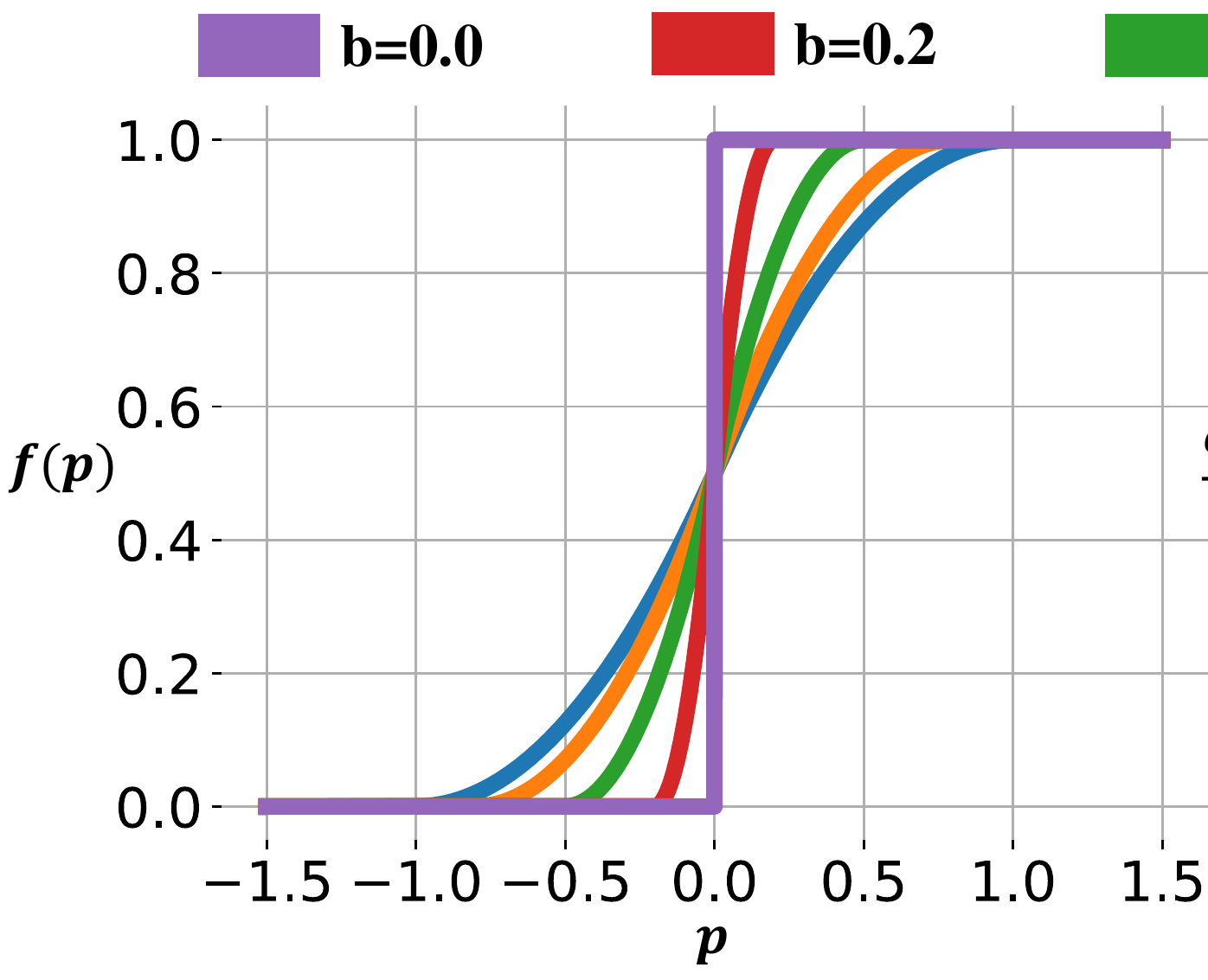}}\subfigure[backward]{\includegraphics[width=0.4\columnwidth]{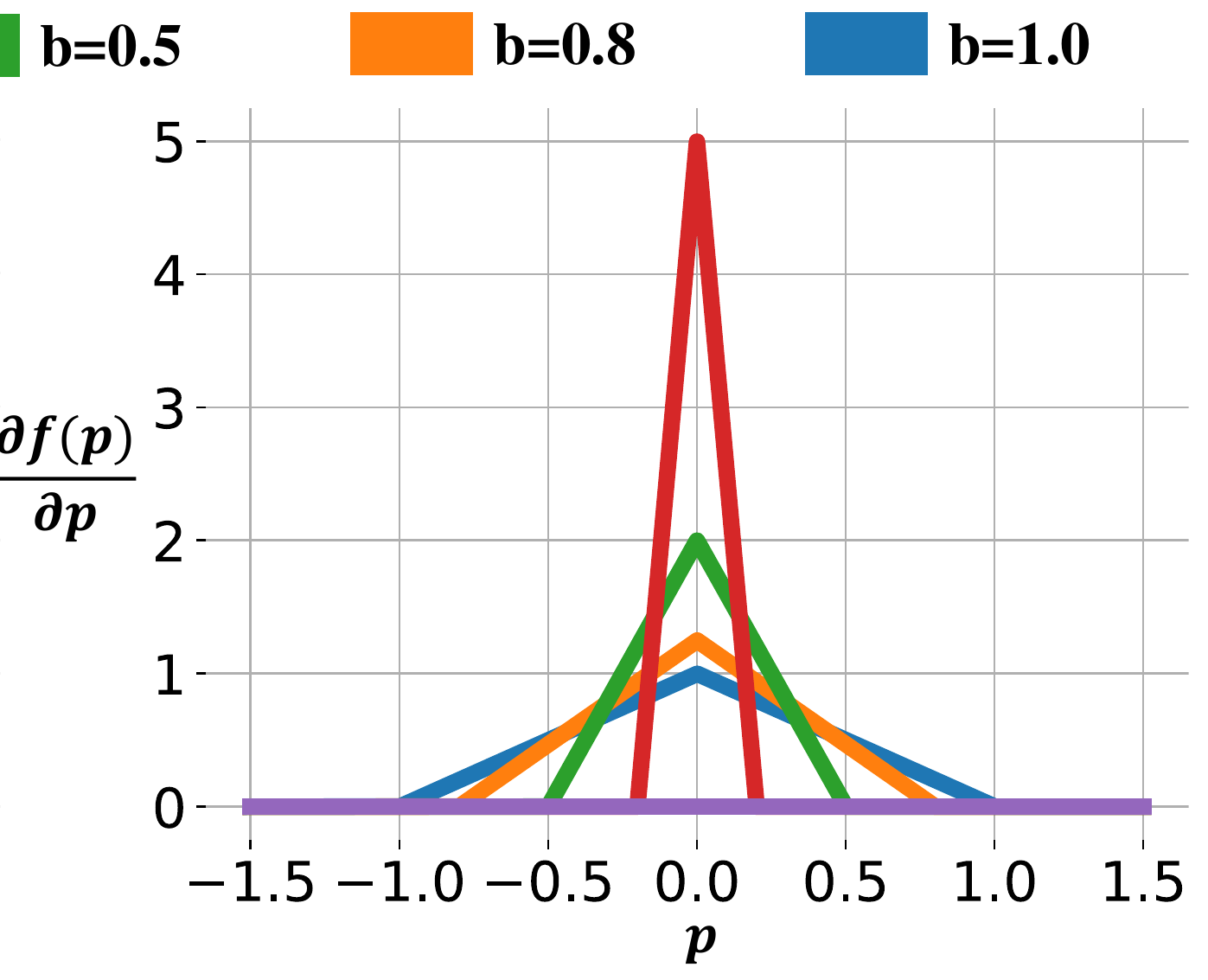}}
\vspace{-0.5em}
\caption{Forward and backward of differentiable mask with varying boundaries.}
\vspace{-1em}
\label{differentiable}
\end{figure}

Inspired by CNN compression, where filter-level sparse constraints are imposed along with the training of CNNs~\cite{liu2017learning,huang2018data,kang2020operation}, we devise a mask set $\mathbf{M} = \{\mathbf{m}_i\}_{i=1}^L$ where $\mathbf{m}_i = \{ m_{i1}, m_{i2}, ..., m_{in_i} \} \in \mathbb{R}^{n_i}$ is the $i$-th layer mask vector. The $j$-th element, \emph{i.e.}, $m_{ij} \in [0, 1]$ will be multiplied with the output of the $j$-th filter of $\mathbf{W}_i^{G_T}$ to scale the convolutional results. Similar to~\cite{liu2017learning,huang2018data}, the network training can be regularized with sparse constraints on the mask $\mathbf{M}$. However, their light-weight network architecture has to be manually defined by a pre-given threshold. In contrast, in this paper, we propose a differentiable mask by regarding each mask $m_{ij} \in \mathbf{m}_i$ as a differentiable function \emph{w.r.t.} a learnable input $p_{ij}$, which in our implementation is designed as:
\begin{equation}\label{mask}
    m_{ij} = f(p_{ij}) = 
    \begin{cases}
     0, & \mbox{if $p_{ij}$ $\le$ $-b$}, \\
     \frac{1}{2} \cdot (\frac{p_{ij} + b}{b})^2, &\mbox{if $p_{ij} \in (-b, 0]$}, \\
     1 - \frac{1}{2} \cdot (\frac{p_{ij} - b}{b})^2, &\mbox{if $p_{ij} \in (0, b)$}, \\
     1, &\mbox{$p_{ij} \ge b$},
    \end{cases}
\end{equation}
where $b$ is a boundary, beyond which, the input $p_{ij}$ will be clipped to $0$ or $+1$. When $p_{ij} \in (-b, 0]$, the mask becomes a convex quadratic function; when $p_{ij} \in (0, b)$, it becomes a concave quadratic function. 
Thus, the interval $(-b, b)$ controls the area of non-step function, and the form of nonlinear function $f(\cdot)$ varies if the boundary $b$ changes as visualized in Fig.\,\ref{differentiable}. In particular, it becomes a step function when $b = 0$.

By changing the boundary value, our mask design is shown to be especially advantageous in its high flexibility. Specifically, a large $b$ can be defined at the training beginning where the quadratic design in Eq.\,(\ref{mask}) will enable the gradient to be updated in a continuous space ($m_{ij} \in [0, 1]$). Then the gradient will gradually be decreased to zero in the end, where Eq.\,(\ref{mask}) will be degenerated to a discrete step function ($m_{ij} \in \{0, 1\}$) if $b$ is set to $0$. The light-weight generator can then be derived by removing zero-masked filters at the test stage. Thus, our mask design closes the gap between the discreteness at test time and the continuity during training.

To further relieve the burden on handcrafted boundary adjustment during training, we devise the following adaptive strategy to keep changing $f(\cdot)$ during the training process:
\begin{equation}\label{boundary}
b = 1 - \sqrt[\uproot{5}3]{e/E},
\end{equation}
where $e$ and $E$ represent the current training iteration and the total number of training iterations ($e \in [0, E]$), respectively. As can be seen, the boundary will start at 1, which leaves a large boundary margin for gradient updating, and end at 0, which denotes the step function. We adopt the third root sign in Eq.\,(\ref{boundary}) to accelerate the convergence of boundary ($1 \rightarrow 0$), since $\sqrt[\uproot{5}3]{
e/E} \ge e/E$ always holds.

\subsubsection{Sparsity Control.}

Our designs in Eq.\,(\ref{mask}) and Eq.\,(\ref{boundary}) can automatically decide which filter to discard (zero-mask) since the introduced parameter $p_{ij}$ is trainable. To obtain the generator with the desired complexity reduction, a sparse mask is essential. To this end, we formulate our sparse regularization loss for the mask set $\mathbf{M}$ as follows:
\begin{equation}\label{sparse}
 \mathcal{L}_{\text{SPA}}={\lambda}_{\text{SPA}}\sum_{i=1}^{L}\sum_{j=1}^{n_i}|p_{ij} + b|, 
\end{equation}
where ${\lambda}_{\text{SPA}}$ is the sparse coefficient.

More specifically, our sparse regularization loss aims to force learnable parameter $p_{ij}$ towards $-b$, which will reward a zero mask according to Eq.\,(\ref{mask}). Different from previous works on CNN compression~\cite{liu2017learning,huang2018data,kang2020operation}, where the exact compression rate can only be obtained after complete training, our training process can be terminated at the point where the sparse rate achieves the desired compression rate, since the zero masks have no gradients and thus will no longer be updated. Thus, the implementation of our sparse design can further reduce the training complexity. Moreover, our sparse constraint is imposed on the learnable parameters rather than the masks directly, which also differentiates our method from existing methods for CNN compression~\cite{liu2017learning,huang2018data,kang2020operation}.

\begin{figure}
\centering
\includegraphics[width=0.75\columnwidth]{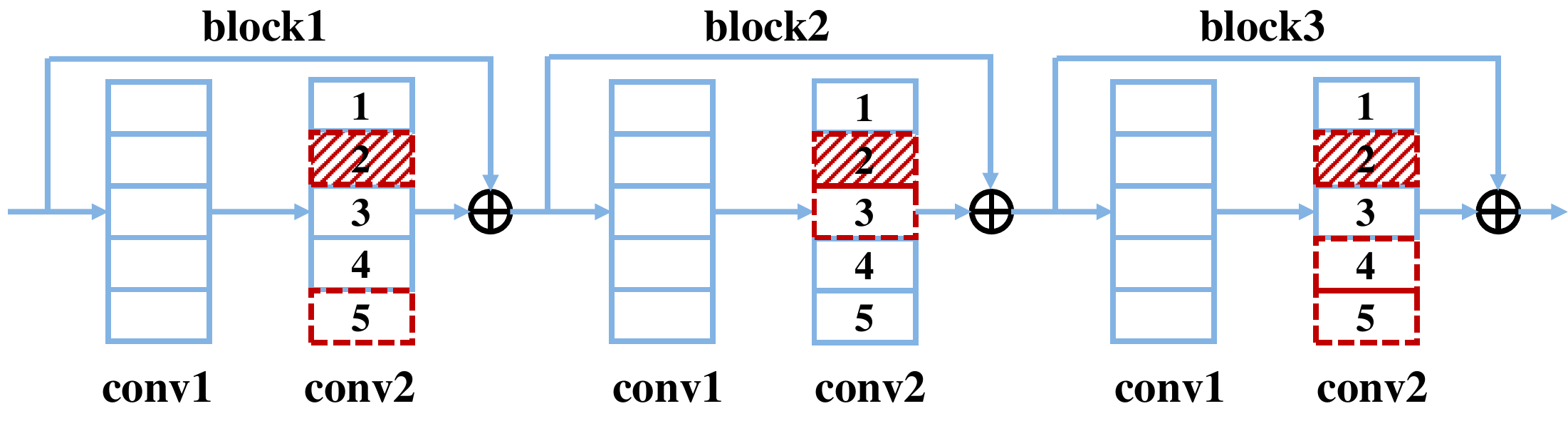}
\vspace{-0.5em}
\caption{An illustration of adaptive cross-block group sparsity. Filters with the same index in the last convolutional layer of all blocks are divided into a group. As can be seen, when the entire group of masks is 0 (red dotted line), the corresponding filters can be removed (like index 2).}
\label{figure:group_sparsity}
\vspace{-1em}
\end{figure}

\subsubsection{Pruning Residual Connections.} 

Residual connections~\cite{he2016deep} are widely adopted in many current neural networks, including CycleGAN~\cite{zhu2017unpaired}, in which the generator network conducts element-wise additions on the feature map outputs of two identity blocks\footnote{The generator of CycleGAN contains a series of identity blocks and the output channel number of all blocks are the same.}. However, pruning the last convolutional layer in one identity block will result in channel inconsistency among different blocks, which paralyzes the entire network.

To solve this, we further propose an adaptive cross-block group sparsity. Specifically, we formulate a set of mask groups $\mathbf{G} = \{\mathbf{g}_1, \mathbf{g}_2, ..., \mathbf{g}_{\tilde{n}} \}$, where $\tilde{n}$ denotes the filter number in the identity block, and each mask group $\mathbf{g}_t = \{m_{1t}, m_{2t}, ..., m_{st} \}$ in which $s$ is the number of identity blocks in the generator and $m_{it}$ is the $t$-th filter in the last convolutional layer of the $i$-th identity block. As shown in Fig.\,\ref{figure:group_sparsity}, the $t$-th mask group consists of the $t$-th filters from all identity blocks. It is easy to know that the channel inconsistency problem can be eliminated by forcing all masks in one group to zeros. To that effect, we devise the following sparse coefficient for the $t$-th mask group:
\begin{equation}\label{group}
    \lambda_{\text{GSPA}_t} =
    \begin{cases}
     0, & \mbox{if $\| \mathbf{g}_t \|_0 = 0$}, \\
     \lambda_{\text{SPA}} \cdot \frac{s}{\| \mathbf{g}_t \|_0}, &\mbox{otherwise},
    \end{cases}
\end{equation}
where $\| \cdot \|_0$ denotes the $\ell_0$-norm which returns the number of non-zero elements of its input. It is intuitive that a higher sparsity of group $\mathbf{g}_t$ returns a larger value of ${\lambda}_{\text{GSPA}_t}$. In each training iteration, we calculate ${\lambda}_{\text{GSPA}_t}$ first, which is then imposed on its corresponding sparsity term in Eq.\,(\ref{sparse}) to replace the coefficient $\lambda_{\text{SPA}}$.

Our motivation of designing Eq.\,(\ref{group}) is to assign a stronger sparse constraint on the mask group with a higher sparsity to push all masks to be zero, such that the corresponding filters in all identity blocks can be safely removed without the problem of channel inconsistency.

\subsection{Co-Attention Distillation}

With the differentiable masks discussed above, we can derive a light-weight generator $G_S$, which is then retrained to restore its performance. However, the performance of $G_S$ is very limited when using traditional adversarial training. According to our experimental observations, the loss of the light-weight generator remains unstable even in the final training iteration, while the discriminator converges to zero quickly. This indicates that $G_S$ can hardly produce a high-quality image to deceive the discriminator. Thus, auxiliary assistance is necessary to stabilize the adversarial training and boost the performance of $G_S$.

Knowledge distillation has emerged as an effective tool in existing GAN compression methods by mimicking intermediate features of a pre-trained teacher GAN in the light-weight generator~\cite{shu2019co,chen2020distilling,fu2020autogan,li2020gan}. However, rather than directly using the intermediate outputs of the teacher model, in this paper, we propose to transfer the attention maps of these intermediate representations, which are demonstrated to contain more valuable information since more details are concentrated there~\cite{zagoruyko2017paying}.

To this end, for each filter weight $\mathbf{W}^{G_S}_i$ in the light-weight generator, following~\cite{zagoruyko2017paying},  we define the attention maps as: 
\begin{equation}\label{student_attention}
 \mathbf{A}^{G_S}_i = \sum_{j=1}^{\tilde{c}_i} | \mathbf{O}^{G_S}_i(j, :, :)|^2,
\end{equation}
where $\mathbf{O}^{G_S}_i$ is the feature map output of $\mathbf{W}^{G_S}_i$. Similarly, we have the attention maps of the teacher generator $\mathbf{A}_i^{G_T} = \sum_{j=1}^{c_i} | \mathbf{O}^{G_T}_i(j, :, :)|^2$.

Our goal is to transfer the attention of the pre-trained teacher model to enrich the learning of the light-weight generator. One simple way to achieve this is to minimize the distance between $\mathbf{A}_i^{G_T}$ and $\mathbf{A}_i^{G_S}$. However, this ignores the efficacy of the teacher discriminator $D_T$, which has a powerful ability to distinguish images of different domains. Inspired by this, we further consider the attention maps from the teacher discriminator, which are then mixed with all the attention maps of the teacher generator to form the co-attention of the teacher:
\begin{equation}\label{teacher_attention}
\mathbf{A}^T_i = \frac{1}{2}(\mathbf{A}_i^{G_T} + \sum_{k=1}^{L'}\mathbf{A}_k^{D_T}),
\end{equation}
%
%where $\mathbf{A}_1^{D_T} = \sum_{j = 1}^{c_1}| \mathbf{O}^{D_T}_1(j, :, :)|^2$ is the attention map of the first convolutional output of the teacher discriminator.
where $\mathbf{A}^{D_T}_k = \sum_{j = 1}^{c_k}| \mathbf{O}^{D_T}_k(j, :, :)|^2$ is the attention map of the teacher discriminator in the $k$-th layer.
To address size inconsistency, we apply Langrange interpolation to the attention map of the discriminator to align it to the attention size of the generator before mixing attentions in Eq.\,(\ref{teacher_attention}). Finally, we derive our objective for co-attention distillation:
\begin{equation}\label{distillation_attention}
    \mathcal{L}_{\text{co-ATT}} = \lambda_{\text{co-ATT}}\sum_{i = 1}^L\big\|\frac{\mathbf{A}^T_i}{\|\mathbf{A}^T_i\|_{2}}-\frac{\mathbf{A}^{G_S}_i}{\|\mathbf{A}^{G_S}_i\|_{2}}\big\|_{2}^{2}.
\end{equation}
where $\lambda_{\text{co-ATT}}$ is a pre-given hyperparameter. We empirically observe that it is not necessary to distill attention from all layers of the teacher network including both the discriminator and generator. Instead, using a small portion can efficiently boost the performance of the light-weight generator, details of which are elaborated in the experiments.

\begin{table*}[!t]
\centering
\caption{Quantitative comparison with other GAN compression methods. $^{\star}$ indicates that a generator with separable convolutions is adopted, following~\cite{li2020gan}. CR stands for the compression rate.}
\begin{tabular}{ccccccc}
\toprule
Model   &Task   &Method   &MACs (CR)   &Parameters (CR)   &FID ($\downarrow$)   &mIoU ($\uparrow$)  \\ \midrule
 \multirow{2}{*}{SAGAN}    & \multirow{2}{*}{CelebA}       & Original & 23.45M (1.0$\times$) & 30.8K (1.0$\times$) & 28.34 & - \\
 & & DMAD (Ours) & 15.46M (1.5$\times$) & 17.3K (1.8$\times$) & 34.97 & - \\ \midrule
 \multirow{2}{*}{StyleGAN2}    & \multirow{2}{*}{CIFAR-10}       & Original &11.54M (1.0$\times$) &2.57M (1.0$\times$) & 37.12 &  \\
 & &DMAD (Ours) &6.12M (1.9$\times$) &1.12M (2.3$\times$) &37.57 &- \\ \midrule
\multirow{31}{*}{CycleGAN} & \multirow{10}{*}{horse2zebra} & Original & 56.8G (1.0$\times$) & 11.3M (1.0$\times$) & 74.04  &  - \\
 & & Co-Evolution~\cite{shu2019co}     & 13.4G (4.2$\times$) & -     & 96.15  &   -   \\
 & & GS~\cite{wang2020gan}             & 12.9G (4.4$\times$) & -     & 86.10  &   -   \\
 & & KD~\cite{aguinaldo2019compressing}& 12.1G (4.7$\times$) & 2.85M (4.0$\times$) & 106.1  &   -   \\
 & & SP~\cite{li2020semantic}          & 12.1G (4.7$\times$) & 2.85M (4.0$\times$) & 86.31  &   -   \\

 & & AutoGAN~\cite{fu2020autogan}      & 6.39G (8.9$\times$) & -     & 83.60   &   -   \\
 & & DMAD (Ours)               & 3.97G (14.3$\times$)  & 0.42M (26.9$\times$)  & 62.41 &   -   \\
 & & GAN-Compression$^{\star}$~\cite{li2020gan} & 2.67G (21.3$\times$) & 0.34M (33.2$\times$) & 64.95 & -   \\
 & & CAT~\cite{jin2021teachers} & 2.55G (22.3$\times$) & - & 60.18 & - \\
 & & DMAD$^{\star}$ (Ours)             & 2.41G (23.6$\times$) & 0.28M (40.4$\times$) & 62.96 &   -   \\ \cline{2-7} 
 & \multirow{7}{*}{zebra2horse} & Original & 56.8G (1.0$\times$) & 11.3M (1.0$\times$)  & 137.8 &    -  \\
 & & Co-Evolution~\cite{shu2019co}     & 13.1G (4.3$\times$) &   -   & 157.9  &   -  \\
 & & GS~\cite{wang2020gan}             & 12.9G (4.4$\times$) &   -   & 120.0  &   -  \\
 & & KD~\cite{aguinaldo2019compressing}& 12.1G (4.7$\times$) & 2.85M (4.0$\times$) & 144.5  &   -   \\
 & & SP~\cite{li2020semantic}          & 12.1G (4.7$\times$) & 2.85M (4.0$\times$) & 140.2  &   -   \\
% }
 & & AutoGAN~\cite{fu2020autogan}      & 4.84G (11.7$\times$) &   -   & 137.2  &   -   \\
 & & DMAD (Ours)               & 3.50G (16.2$\times$)  & 0.30M (37.7$\times$) &  139.3  &   -   \\ \cline{2-7} 
 & \multirow{7}{*}{summer2winter} & Original & 56.8G (1.0$\times$) & 11.3M (1.0$\times$)  & 79.12  &   -   \\
 & & Co-Evolution~\cite{shu2019co}     & 11.1G (5.1$\times$)&   -   & 78.58  &   -   \\
 & & GS~\cite{wang2020gan}             & 9.15G (6.2$\times$)&   -   & 70.21  &   -   \\
 & & AutoGAN~\cite{fu2020autogan}      & 4.34G (13.1$\times$) &   -   & 78.33  &   -   \\
 & & KD~\cite{aguinaldo2019compressing}& 3.20G (17.8$\times$) & 0.72M (15.7$\times$) & 80.10  &   -   \\
 & & SP~\cite{li2020semantic}          & 3.20G (17.8$\times$) & 0.72M (15.7$\times$) & 76.59  &   -   \\
 & & DMAD (Ours)               & 3.18G (17.9$\times$) & 0.24M (47.1$\times$) &  78.24  &   -   \\ \cline{2-7} 
 & \multirow{7}{*}{winter2summer} & Original & 56.8G (1.0$\times$) & 11.3M (1.0$\times$)  & 73.31  &   -   \\
 & & Co-Evolution~\cite{shu2019co}     & 11.0G (5.2$\times$) &  -   & 79.16  &   -   \\
 & & GS~\cite{wang2020gan}             & 9.45G (6.0$\times$)&   -   & 74.80  &   -   \\
 & & AutoGAN~\cite{fu2020autogan}      & 4.26G (13.3$\times$)  &  -   & 77.73  &   -   \\
& & KD~\cite{aguinaldo2019compressing}& 3.20G (17.8$\times$) & 0.72M (15.7$\times$) & 80.10  &   -   \\
 & & SP~\cite{li2020semantic}          & 3.20G (17.8$\times$) & 0.72M (15.7$\times$) & 76.59  &   -   \\
 & & DMAD (Ours)               & 4.29G (13.2$\times$)  & 0.45M (25.1$\times$) &  70.97  &   -   \\ \midrule
\multirow{15}{*}{Pix2Pix} & \multirow{6}{*}{edges2shoes} & Original & 18.6G (1.0$\times$) & 54.4M (1.0$\times$) & 34.31  & - \\
 & & Pix2Pix 0.5$\times$   & 4.65G (4.0$\times$)  & 13.6M (4.0$\times$)& 52.02  &   -   \\

 & & DMAD (Ours)               & 2.99G (6.2$\times$)  & 2.13M (25.5$\times$) & 46.95 &   - \\
 & & GAN-Compression$^{\star}$~\cite{li2020gan} & 4.81G (3.8$\times$)   & 0.7M (77.7$\times$)  &  26.60   &   -   \\
  & & PD~\cite{gong2021towards}*  & 4.56G (4.1$\times$)  & 0.51M(106.7$\times$) &  25.96  &   -   \\
 & & DMAD$^{\star}$ (Ours)             & 4.30G (4.3$\times$)  & 0.54M (100.7$\times$) &  24.08  &   -   \\ \cline{2-7} 
 & \multirow{9}{*}{cityscapes} & Original & 18.6G (1.0$\times$) & 54.4M (1.0$\times$)  &  - & 42.71 \\
 & & Slimming~\cite{liu2017learning} & 5.01G (3.7$\times$) & 1.22M (44.2$\times$) & - & 37.33\\
 & & Pix2Pix 0.5$\times$                & 4.65G (4.0$\times$)  & 13.6M (4.0$\times$)&    -   & 39.02 \\
 %Params: 1.85M | MACs: 4116.52M mIOU:38.69
 & & SSS~\cite{huang2018data} & 4.12G (4.5$\times$)  & 1.85M (29.4$\times$) & - & 38.69 \\
 & & DMAD (Ours)              & 3.96G (4.7$\times$)  & 1.73M (31.4$\times$) & - &    40.53  \\
 & & GAN-Compression$^{\star}$~\cite{li2020gan} & 5.66G (3.3$\times$) & 0.71M (76.6$\times$) &    -   & 40.77 \\
 & & PD~\cite{gong2021towards}*  & 3.69G (5.0$\times$)  & 0.58M(93.8$\times$) &  -  &   35.03   \\
  & & CAT~\cite{jin2021teachers}* & 5.57G (3.3$\times$) & - & - & 42.53 \\
 & & DMAD$^{\star}$ (Ours)             & 4.39G (4.2$\times$) & 0.55M (98.9$\times$) &  -  &   41.47\\ \bottomrule
\end{tabular}
\label{tab:quantitative_evaluation}
\end{table*}

In addition to the proposed co-attention distillation, the high-level information outputs of discriminator are also crucial. Following~\cite{chen2020distilling}, we directly match the last convolutional outputs of the discriminator for the images generated by the pre-trained and light-weight generators as:
\begin{equation}\label{FEA}
    \mathcal{L}_{\text{FEA}}=\lambda_{\text{FEA}}\dfrac{1}{n}\sum_{i=1}^{n}\big\|\tilde{D}_{T}\left(G_{T}\left(x_{i}\right)\right)-\tilde{D}_{T}\left(G_{S}\left(x_{i}\right)\right)\big\|_{2}^{2},
\end{equation}
where $\lambda_{\text{FEA}}$ is a pre-defined hyperparameter and $\tilde{D}_{T}$ returns the feature map outputs from the last convolutional layer of the pre-trained discriminator. Then, our final adversarial training objective can be obtained as:
\begin{equation}
\mathcal{L} = \mathcal{L}_{\text{cGAN}} + \mathcal{L}_{\text{SPA}} + \mathcal{L}_{\text{co-ATT}} + \mathcal{L}_{\text{FEA}}.
\end{equation}

% Lastly, it is easy to derive the derivative of $\mathcal{L}$ \emph{w.r.t.} the learnable parameter $p_{ij}$ in the differentiable mask as

% \begin{equation}
%     \frac{\partial \mathcal{L}}{\partial p_{ij}} = 
%     \begin{cases}
%      1 + \frac{p_{ij} + b}{b^2}, &\mbox{if $p_{ij} \in (-b, 0]$}, \\
%      1 - \frac{p_{ij} - b}{b^2}, &\mbox{if $p_{ij} \in (0, b)$}, \\
%      0, &\mbox{otherwise}.
%     \end{cases}
% \end{equation}

\section{Experiments}\label{experiment}

\subsection{Experimental Settings}

\subsubsection{Model and Datasets.} 

Following~\cite{shu2019co,chen2020distilling,fu2020autogan,li2020gan}, we conduct experiments using CycleGAN~\cite{zhu2017unpaired} and Pix2Pix~\cite{isola2017image}. CycleGAN was devised for unpaired image-to-image translation, and is thus evaluated on unpaired datasets, including horse2zebra~\cite{zhu2017unpaired} and summer2winter~\cite{zhu2017unpaired}. In contrast, Pix2Pix is a paired image-to-image translation network; thus, we evaluate it on paired datasets, including edges2shoes~\cite{yu2014fine} and Cityscapes~\cite{cordts2016cityscapes}. Besides, following~\cite{li2020gan}, where the generators are replaced with a separable convolutional network, we also conduct experiments using such a generator structure for fair comparison.
In addition, we apply DMAD to SAGAN~\cite{zhang2019self} on CelebA~\cite{liu2015deep} to show the applicability of our DMAD on unconditional GANs. Besides, we also conduct experiments using the popular StyleGAN2~\cite{karras2020analyzing} on CIFAR-10~\cite{krizhevsky2009learning} dataset.

\subsubsection{Evaluation Metrics.} Following previous works for GAN compression~\cite{shu2019co,fu2020autogan,li2020gan}, we use FID (Frechet Inception Distance)~\cite{heusel2017gans} to measure the similarity between real images and generated images. A lower FID score denotes a higher quality of generated images. On cityscapes, we adopt the DRN-D-105~\cite{yu2017dilated} segmentation model to calculate the mIoU (mean Intersection over Union) for evaluating the quality of generated images. Different from FID, a larger value of mIoU implies a better generated images. Besides, we also report the MACs, parameters and their compression rates for comparison.

\subsubsection{Implementation Details. \label{implementation_details}} 

For all experiments, we adopt the Adam optimizer with a tuple of beta values as (0.5,0.999) for both generator and discriminator. The learning rate is kept to 0.0002 in the beginning and linearly decayed to zero. Following the same settings in~\cite{zhu2017unpaired,isola2017image}, the batch sizes on horse2zebra, summer2winter, edges2shoes and cityscapes are set to 1, 1, 4, and 1, respectively. The hyperparameters ${\lambda}_{\text{SPE}}$, ${\lambda}_{\text{SPA}}$, ${\lambda}_{\text{co-ATT}}$ and ${\lambda}_{\text{FEA}}$ in this paper are set to 10, 0.001, 100, 0.0001 for horse2zebra and summer2winter, and 100, 0.01, 50, 0.0001 for edges2shoes and cityscapes, respectively. As for SAGAN and StyleGAN2, we set ${\lambda}_{\text{SPA}}$, ${\lambda}_{\text{co-ATT}}$ and ${\lambda}_{\text{FEA}}$ to 0.001, 10, 0.0001, respectively.

In terms of co-attention distillation, we split the CycleGAN generator into five equal groups (each residual block is regarded as one layer), and then select per-group feature map outputs to extract attention maps. As for the Pix2Pix generator, we extract attention maps from the second, fourth, twelfth, and fourteenth convolutional layers.

\begin{figure*}[!t]
\begin{center}
\includegraphics[width=0.7\linewidth]{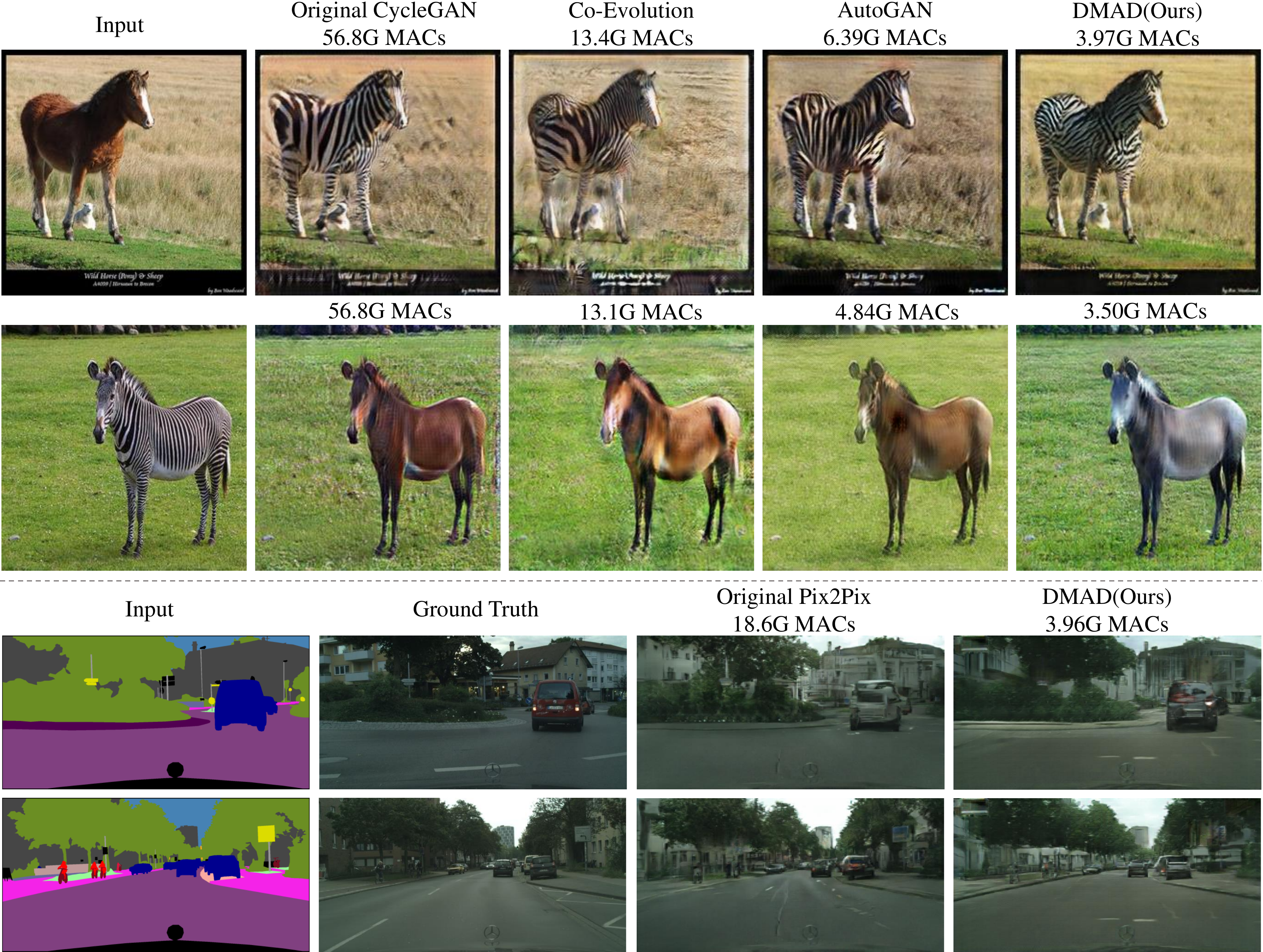}
\end{center}
\vspace{-1em}
\caption{Visualization of our DMAD. The first two lines are the results on horse2zebra and zebra2horse, comparing with Co-Evolution~\cite{shu2019co} and AutoGAN~\cite{fu2020autogan}. The third line displays the results on cityscapes.}
\vspace{-1em}
\label{figure:visualization_results}
\end{figure*}

\subsection{Experimental Results}

\subsubsection{Quantitative Results.} 

As can be seen from Tab.\,\ref{tab:quantitative_evaluation}, for CycleGAN compression, our DMAD can achieve a comparable (on zebra2horse, summer2winter and winter2summer) or even a much smaller (on horse2zebra) FID metric in comparison to the original model, while still obtaining a significant reduction in model complexity of 13.2 -- 23.6$\times$ in MACs and 25.1 -- 47.1$\times$ in parameters. Similarly, for Pix2Pix compression, the proposed DMAD results in 4.24 -- 6.22$\times$ MACs compression and 25.5 -- 100.7$\times$ reductions of parameters, while obtaining a smaller FID value on edges2shoes and similar mIOU performance on cityscapes when compared to the full model. 
Besides, in contrast to the recent state-of-the-arts~\cite{shu2019co,fu2020autogan,li2020gan,jin2021teachers}, our compressed models have fewer MACs and parameters, while still retaining a better metric performance in most cases. This demonstrates the effectiveness of our DMAD in reducing GAN complexity. Besides, comparing to the slimming Pix2Pix 0.5$\times$, that is, the direct use of a scale factor, our differentiable mask produces a non-uniform architecture structure which leads to better performance. This result is consistent with the recent studies on network pruning~\cite{liu2018rethinking,lin2020channel}. Compared with other differentiable mask methods~\cite{liu2017learning, huang2018data}, our proposed DMAD also achieves superior performance than Slimming~\cite{liu2017learning} and SSS~\cite{huang2018data} on cityscapes dataset. Moreover, from Tab.\,\ref{tab:quantitative_evaluation}, we can also observe that when compressing SAGAN on CelebA and StyleGAN2 on CIFAR-10, DMAD can show its excellent compression ability as well, even although these models are already computationally tiny on these datasets.

\begin{table}[!t]
\centering
\caption{Ablation experiment of mask function in our proposed differentiable mask. The experiments are conducted with CycleGAN on horse2zebra and zebra2horse.}
\begin{tabular}{ccccc}
\toprule
Method                       & Task        & MACs  & Parameters & FID ($\downarrow$)   \\ \midrule
\multirow{2}{*}{Linear Mask} & horse2zebra & 4.43G & 0.36M      & 85.74 \\ \cline{2-5} 
                             & zebra2horse & 4.83G & 0.53M      & 145.9 \\ \midrule
\multirow{2}{*}{DMAD (Ours)}        & horse2zebra & 3.97G & 0.42M      & 62.41 \\ \cline{2-5} 
                             & zebra2horse & 3.50G & 0.30M      & 139.3 \\ \bottomrule
\end{tabular}
\label{tab:mask_function}
\end{table}

\begin{table}[!t]
\caption{Performance comparison with/without our adaptive cross-block group sparsity on horse2zebra and zebra2horse. PR represents the pruning rate of residual blocks.}
% \resizebox{\columnwidth}{!}{
\centering
\begin{tabular}{ccccc}
\toprule
Task    & Adaptive    & PR & MACs  & FID ($\downarrow$)     \\ \midrule
\multirow{2}{*}{horse2zebra} &  $\times$   &    0.33      & 6.80G &   70.77 \\ 
                                         & \checkmark  &    0.72      & 3.97G &   66.23 \\ \midrule
\multirow{2}{*}{zebra2horse} &  $\times$   &    0.39      & 5.43G &   147.7 \\ 
                                         & \checkmark  &    0.69      & 3.50G &   140.2 \\ \bottomrule
\end{tabular}
% }
\label{tab:adaptive_pruning}
\end{table}

% \subsubsection{Adaptive Cross-Block Group Sparsity.}

% Tab.\,\ref{tab:adaptive_pruning} displays the effectiveness of our adaptive cross-block group sparsity for residual connections in CycleGAN. We remove the co-attention distillation to exclude its effect. We can see that our adaptive cross-block group sparsity can effectively reduce the model redundancy with better image generation. Thus, this clearly demonstrates the necessity of pruning residual connections and efficacy of our adaptive sparsity strategy.
\subsubsection{Differentiable Mask.} We verify the efficacy of our differentiable mask from two aspects: the design of our quadratic mask function and the design of our adaptive cross-block group sparsity.

In Tab.\,\ref{tab:mask_function}, we replace our quadratic mask with the linear version in~\cite{huang2018data, liu2017learning}. As can be seen, with a better ability to reduce more complexity in computation and overload, our quadratic mask still results a significantly better FID performance, well demonstrating the advantage of our quadratic design over the linearity.

Tab.\,\ref{tab:adaptive_pruning} displays the effectiveness of our adaptive cross-block group sparsity for residual connections in CycleGAN. We remove the co-attention distillation to exclude its effect. We can see that our adaptive cross-block group sparsity can effectively reduce the model redundancy with better image generation. This demonstrates the necessity of pruning residual connections and efficacy of our adaptive sparsity strategy.

\subsubsection{Component Analysis.} 

In Tab.\,\ref{tab:ablation_study}, we show the impact of different components in our DMAD including inheriting weights from the searched model, generator attention, our proposed co-attention, and high-level information of discriminator. As can be seen from the table, inheriting weights (Finetune) from the pre-trained model does benefit to the quality of generated images. Then, by considering the generator attention (G-Attention), the FID metric is further reduced which implicates the effectiveness of distilling attention mechanism in improving the performance of light-weight generator. Besides, with attention maps from the discriminator (co-Attention), the generated images are further improved with FID decreasing from 66.26 to 64.97. Such a performance gain verifies the efficacy of our proposed co-attention distillation. Moreover, the distilling high-level information of discriminator (D-Distill) also benefits the generated images where the FID decreases from 67.26 to 65.38. By utilizing all of these components, our DMAD obtains the best FID value of 62.41.

% When inheriting the weights of the search process, the performance is better than when training from scratch. The weights obtained during the search process provide a warm-up for retraining, and a good initialization for the network. Therefore, our experiments directly adopt the strategy of inheriting weights. In addition, both $\mathcal{L}_{\text{co-ATT}}$ and $\mathcal{L}_{\text{FEA}}$ can improve the performance of the model, and combining them achieves the best results.

\begin{table}[!t]
\centering
\caption{The impact of different components on horse2zebra tasks. G-Attention denotes that only generator attentions are applied. D-Distill denotes distillation of high-level information from discriminator.}
\begin{tabular}{ccccc}
\toprule
Model              & Inherit Weights   & $\mathcal{L}_{\text{co-ATT}}$ & $\mathcal{L}_{\text{FEA}}$ & FID ($\downarrow$)       \\ \midrule
Scratch            &                   &                      &                          &   69.50  \\
Finetune           & \checkmark        &                      &                          &   67.26  \\
G-Attention     & \checkmark        &                      &                          &   66.26  \\
co-Attetnion       & \checkmark        &     \checkmark       &                          &   64.97  \\
D-Distill      & \checkmark        &                      &        \checkmark        &   65.38  \\
DMAD (Ours)         & \checkmark        &     \checkmark       &        \checkmark        &   62.41  \\ \bottomrule
\end{tabular}
\label{tab:ablation_study}
\end{table}

\subsubsection{Visualization.}

Fig.\ref{figure:visualization_results} shows the visualization results on horse2zebra, zebra2horse, and cityscapes. As can be seen, our method obtains similar or even better visual results, with higher MACs reductions. Compared with AutoGAN~\cite{fu2020autogan}, where the zebra stripes very sharp, and Co-evoluation~\cite{shu2019co}, where clear softening marks are generated around the horse, our generated images are more vivid and authentic. On cityscapes, we compress the Pix2Pix generator by 4.7$\times$, while still achieving a similar output as the original generator.

\begin{table}[]
\centering
\caption{Comparison of different attention map fusion scenarios in co-attention distillation.}
\begin{tabular}{ccc}
\toprule
Task                         & Fusion Method   & FID ($\downarrow$) \\ \midrule
\multirow{3}{*}{horse2zebra} & multiplication  & 71.05  \\ \cline{2-3} 
                             & 1$\times$1 convolution & 73.45  \\ \cline{2-3} 
                             & summation (Ours) & 62.41  \\ \midrule
\multirow{3}{*}{zebra2horse} & multiplication  & 140.5  \\ \cline{2-3} 
                             & 1$\times$1 convolution & 141.3  \\ \cline{2-3} 
                             & summation (Ours) & 139.3  \\ \bottomrule
\end{tabular}
\label{table:fusion_methods}
\end{table}

\subsubsection{Fusion Methods.}
In Eq.\,(\ref{student_attention}), we define the attention map by summing up the feature map outputs.
Tab.\,\ref{table:fusion_methods} further the comparison between our summation and other two scenarios including multiplication and 1$\times$1 convolution. The multiplication directly multiplies the attention map between generator and discriminator. Although highlighting the overlapping parts, it also ignores the individual specific activation regions. Therefore, the multiplication results in poor performance. Similarly,  1$\times$1 convolution cannot guarantee the quality of the fusion due to the lack of suitable supervision signals. Consequently, it leads to a poor distillation performance. In contrast, the summation of the attention maps can highlight the overlapping parts between the attention maps and maintain their specific activation regions. Despite its simpleness, it is of great effectiveness.

\begin{table}[!t]
\centering
\caption{The impact of total distillation layers in teacher discriminator. We compress Pix2Pix on cityscapes dataset.}
\begin{tabular}{c|cccc}
\toprule
Total Distillation Layers & 1     & 2     & 3     & 4     \\ \midrule
mIOU      & 40.53 & 37.33 & 37.18 & 39.09 \\ \bottomrule
\end{tabular}
\label{table:various_discriminator_layers}
\end{table}

\subsubsection{No. of Distillation layers.}
We follow GAN Compression~\cite{li2020gan} to select the distillation layer of the teacher generator. By distilling features in every three layers, GAN Compression~\cite{li2020gan} has verified that this manner can transfer enough knowledge for learning a light-weight student generator.

We further explore the effect of distilling more layers of the discriminator's characteristics.
The results are shown in Tab.~\ref{table:various_discriminator_layers}. 
As can be seen, considering only one layer brings about the best performance. In our implementation, we only extract the layer whose output feature size is similar to the size of the feature extracted by the light-weight generator. To dive into a deeper analysis, different feature maps are in different sizes, thus the size alignment is required to the purpose of distillation. Though different layers might contain different information, the size alignment destroys the information, which might cause the poor performance.

\begin{figure}[!t]
\begin{center}
\includegraphics[width=0.8\columnwidth]{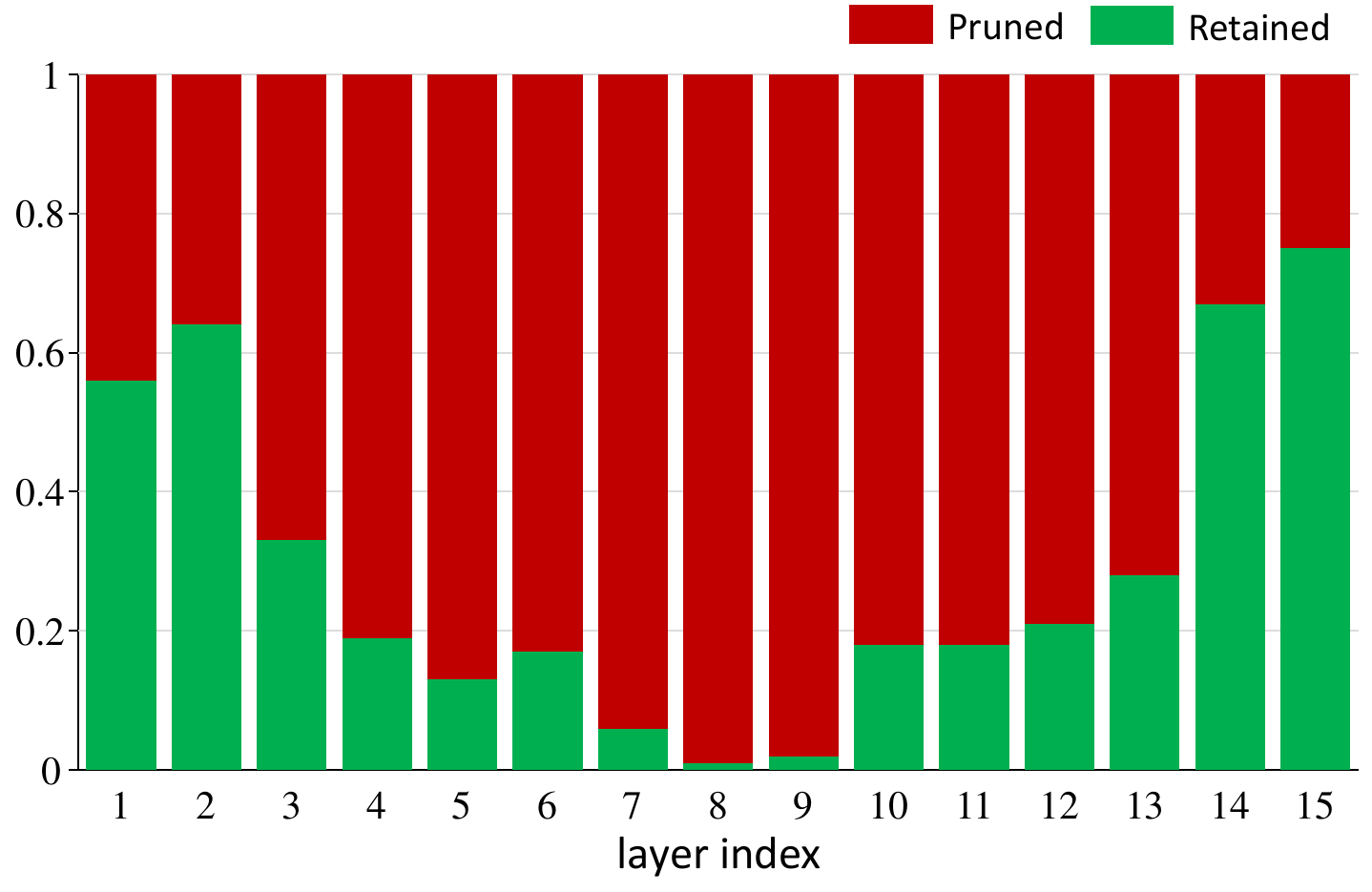}
\end{center}
\vspace{-1em}
\caption{Per-layer pruning rate of Pix2Pix generator on cityscapes.}
\vspace{-1em}
\label{figure:pruning_rate_unet}
\end{figure}

\subsubsection{Compression and Acceleration.} The pruning rates of each layer searched by the differential mask in Pix2Pix on the cityscapes dataset as shown in Fig.\ref{figure:pruning_rate_unet}. Pix2Pix uses U-Net as generator, we find that, as the number of U-Net network layers increases, the overall pruning rate looks like an letter U. The first few layers extract valid input information, while the last few layers construct the final generated images, so these layers are vital for generating high-quality images. Compared to these layers, the other layers of U-Net are less important. This is especially true for the most intermediate network layer, whose output feature map size is only 1$\times$1, and has little impact on the final 256$\times$256 image. Thus, these layers can reach a higher pruning rate. The results reflect that the differential mask can adaptively prune according to the contribution of each layer.

During model deployment, the actual acceleration effect on the hardware is more critical than the reduction of theoretical calculations. We compare the actual acceleration effect between the original model and the compressed model on different types of processors (\emph{i.e.}, CPUs and GPUs) under Pytorch V1.5.1. On the horse2zebra task, it takes 440ms for the original model to process one image using an Intel Xeon CPU E5-2690 CPU and 7.1ms on one GTX1080TI GPU. The compressed model requires 76.3ms and 5.8ms to process an image on a CPU and a GPU, respectively. It achieves 5.8$\times$ and 1.2$\times$ acceleration on a CPU and a GPU, respectively. Due to the powerful parallel processing capabilities of GPUs, the acceleration effect on the GPU is not as obvious as on the CPU. The gap between the realistic and theoretical speed-up ratio may come from the limitations of IO decay, data transmission, \emph{etc}.

\begin{table}[]
\centering
\caption{The compressed generator is trained with different discriminator network widths. The experiments are conducted using Pix2Pix on cityscapes dataset. The original discriminator width is 128.}
\begin{tabular}{c|cccc}
\toprule
network width of discriminator & 128   & 64    & 32    & 16    \\ \midrule
mIOU                         & 38.94 & 39.88 & 37.39 & 36.64 \\ \bottomrule
\end{tabular}
\label{table:diff_ndf}
\end{table}

\subsubsection{Adversarial Training Balance.} Simply compressing the generator breaks the Nash equilibrium between the generator and the discriminator during adversarial training, causing poor performance of the compressed generator. We conduct experiments using Pix2Pix on the cityscapes dataset. We train the compressed generator in Pix2Pix with different network widths of the discriminator to retain the Nash equilibrium as much as possible. The experimental results in the Tab.\,\ref{table:diff_ndf} show that the network width of the discriminator does have an impact on the performance of the compression generator.
When the width is 64, the best mIOU of 39.88 can be obtained. Nevertheless, the improvement is limited in comparison with 40.53 mIOU from our proposed distillation as shown in Tab.\,\ref{tab:quantitative_evaluation}.

\section{Conclusion}
In this work, we introduce a novel framework for GAN compression, termed DMAD. First, we propose a differentiable mask to carry out architecture search for a light-weight generator in a training-adaptive manner. An adaptive cross-block group sparsity is then further incorporated for block-level pruning. Second, we distill attention maps extracted from both the generator and discriminator of the pre-trained model to the light-weight generator, which makes training more stable and yields superior performance. Extensive experiments demonstrate the ability of our proposed DMAD method in preserving visual quality of generated images.

% or



% \section*{Acknowledgment}


\ifCLASSOPTIONcaptionsoff
  \newpage
\fi



\bibliographystyle{IEEEtran}
\bibliography{reference}

\begin{IEEEbiography}[{\includegraphics[width=1in,height=1.25in,clip,keepaspectratio]{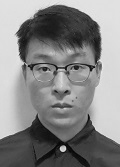}}]{Shaojie Li}
studied for his B.S. degrees in FuZhou University, China, in 2019. He is currently trying to pursue a M.S. degree in Xiamen University, China. His research interests include model compression and computer vision.
\end{IEEEbiography}

\begin{IEEEbiography}[{\includegraphics[width=1in,height=1.25in,clip,keepaspectratio]{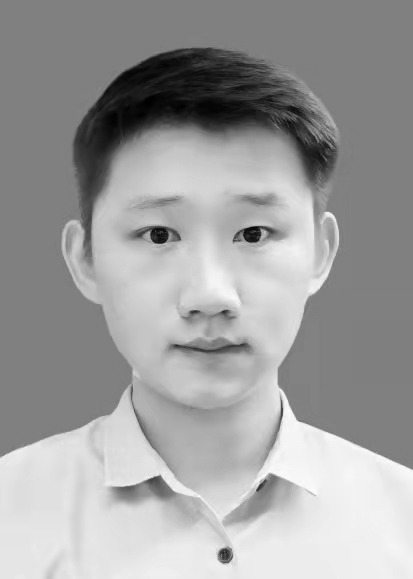}}]{Mingbao Lin} is currently pursuing the Ph.D degree with Xiamen University, China. He has published over ten papers as the first author in international journals and conferences, including IEEE TPAMI, IJCV, IEEE TIP, IEEE TNNLS, IEEE CVPR, NeurIPS, AAAI, IJCAI, ACM MM and so on. His current research interest includes network compression \& acceleration, and information retrieval.
\end{IEEEbiography}

\begin{IEEEbiography}[{\includegraphics[width=1in,height=1.25in,clip,keepaspectratio]{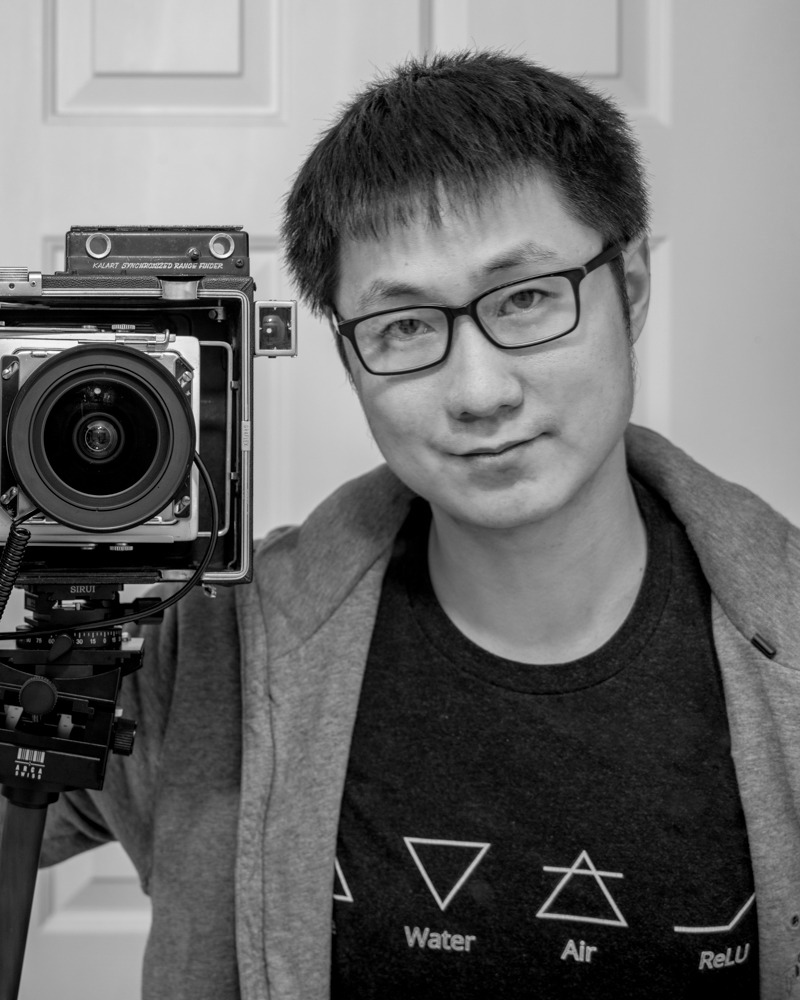}}]{Yan Wang}
works as a software engineer in Search at Pinterest. With a Ph.D degree on Electrical Engineering from Columbia University, Yan published over 20 papers on top international conferences and journals, and holds 10 US or international patents. He has broad interests on deep learning's applications on multimedia retrieval.
\end{IEEEbiography}

\begin{IEEEbiography}[{\includegraphics[width=1in,height=1.25in,clip,keepaspectratio]{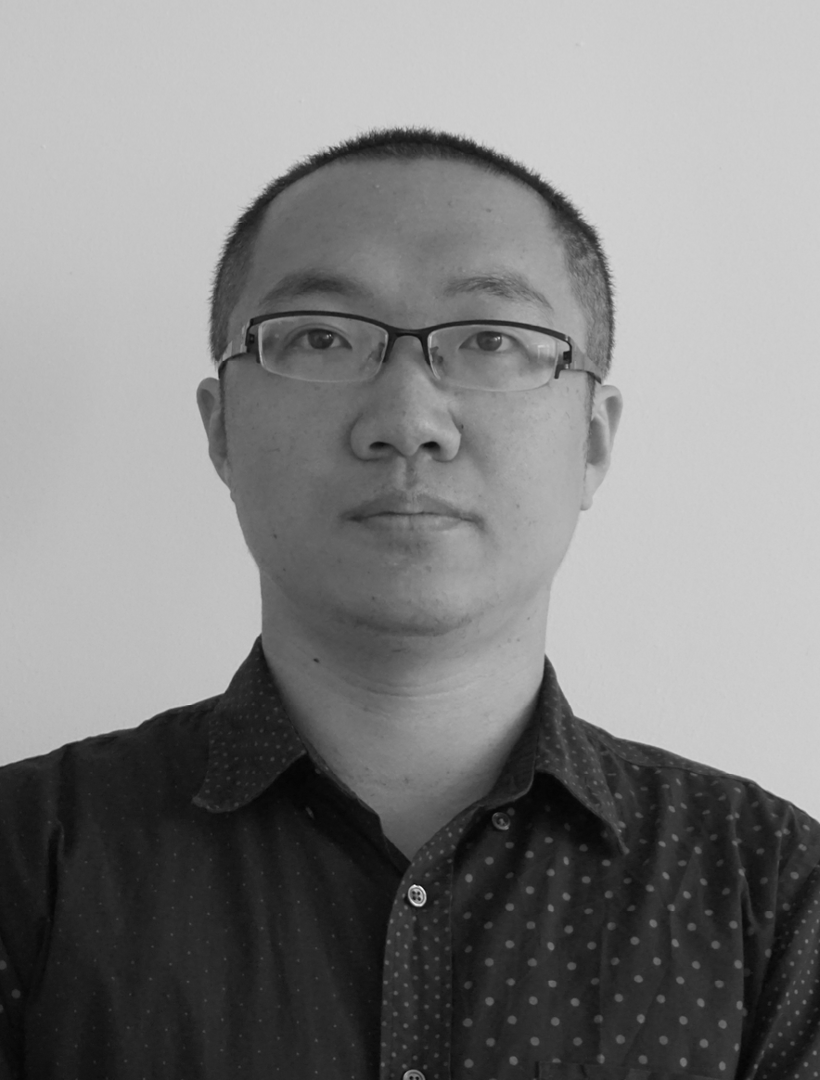}}]{Fei Chao} (M'11) received the B.Sc. degree in Mechanical Engineering from the Fuzhou University, China, and the M.Sc. Degree with distinction in Computer Science from the University of Wales, Aberystwyth, U.K., in 2004 and 2005, respectively, and the Ph.D. degree in robotics from the Aberystwyth University, Wales, U.K. in 2009. He is currently an Associate Professor with the School of Informatics, Xiamen University, China. Dr Chao has published more than 100 peer-reviewed journal and conference papers. His research interests include developmental robotics, machine learning, and optimization algorithms.
\end{IEEEbiography}

% \begin{IEEEbiography}[{\includegraphics[width=1in,height=1.25in,clip,keepaspectratio]{xdm.png}}]{Xudong Mao} is currently an associate professor at Xiamen University. Before joining XMU, he was a postdoctoral fellow at The Hong Kong Polytechnic University. He received the PhD degree in Computer Science from City University of Hong Kong, and received the BSc degree in Information Security from Nankai University, China. His research interests are in the areas of generative adversarial networks, image-to-image translation, and computer vision.
% \end{IEEEbiography}

% \begin{IEEEbiography}[{\includegraphics[width=1in,height=1.25in,clip,keepaspectratio]{mlx.png}}]{Mingliang Xu} received the Ph.D. degree in com- puter science and technology from the State Key Lab of CAD and CG, Zhejiang University, Hangzhou, China, in 2011. He is currently a Full Professor with the School of Information Engineering, Zhengzhou University, Zhengzhou, China, where he is currently the Director of the Center for Interdisciplinary Information Science Research and the Vice General Secretary of ACM SIGAI China. His current research inter- ests include computer graphics, multimedia, and artificial intelligence.
% \end{IEEEbiography}

\begin{IEEEbiography}[{\includegraphics[width=1in,height=1.25in,clip,keepaspectratio]{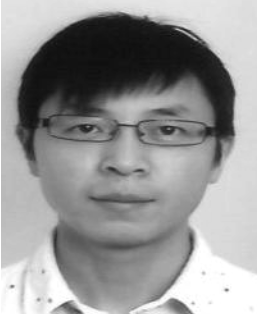}}]{Ling Shao} (Fellow, IEEE) is currently the Executive Vice President and a Provost of the Mohamed bin Zayed University of Artificial Intelligence. He is also the CEO and the Chief Scientist of the Inception Institute of Artificial Intelligence (IIAI), Abu Dhabi, United Arab Emirates. His research interests include computer vision, machine learning, and medical imaging. He is a fellow of IEEE, IAPR, IET, and BCS.
\end{IEEEbiography}

\begin{IEEEbiography}[{\includegraphics[width=1in,height=1.25in,clip,keepaspectratio]{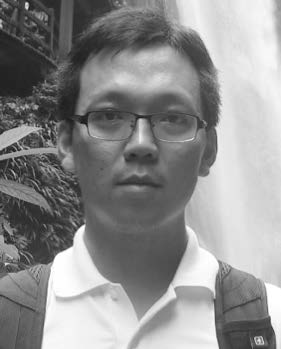}}]{Rongrong Ji}
(Senior Member, IEEE) is a Nanqiang Distinguished Professor at Xiamen University, the Deputy Director of the Office of Science and Technology at Xiamen University, and the Director of Media Analytics and Computing Lab. He was awarded as the National Science Foundation for Excellent Young Scholars (2014), the National Ten Thousand Plan for Young Top Talents (2017), and the National Science Foundation for Distinguished Young Scholars (2020). His research falls in the field of computer vision, multimedia analysis, and machine learning. He has published 50+ papers in ACM/IEEE Transactions, including TPAMI and IJCV, and 100+ full papers on top-tier conferences, such as CVPR and NeurIPS. His publications have got over 10K citations in Google Scholar. He was the recipient of the Best Paper Award of ACM Multimedia 2011. He has served as Area Chairs in top-tier conferences such as CVPR and ACM Multimedia. He is also an Advisory Member for Artificial Intelligence Construction in the Electronic Information Education Committee of the National Ministry of Education.
\end{IEEEbiography}

\end{document}